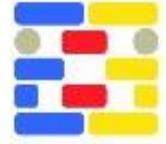

# LLM-assisted Graph-RAG Information Extraction from IFC Data

Sima Iranmanesh*, Hadeel Saadany*, and Edlira Vakaj*
* College of Computing, Birmingham City University, birmingham, UK

**Abstract**

IFC data has become the general building information standard for collaborative work in the construction industry. However, IFC data can be very complicated because it allows for multiple ways to represent the same product information. In this research, we utilise the capabilities of LLMs to parse the IFC data with Graph Retrieval-Augmented Generation (Graph-RAG) technique to retrieve building object properties and their relations. We will show that, despite limitations due to the complex hierarchy of the IFC data, the Graph-RAG parsing enhances generative LLMs like GPT-4o with graph-based knowledge, enabling natural language query-response retrieval without the need for a complex pipeline.

## Introduction

BIM (Building Information Modelling) technology is used in the architecture, engineering, construction, and operation (AECO) industry to manage and share building information throughout the construction cycle. Since buildingSMART (Liebich et al., 2010) has published the IFC (Industry Foundation Classes), the BIM servers generally use IFC data as a standard for storing, exchanging and extracting information on architectural elements (Xu et al., 2018; Faraj et al., 2000). IFC files are essentially information-intensive complex text files that share data throughout the project life cycle across disciplines and technical applications. There are various commercial as well as open source tools available for working with IFC data. However, extracting information from IFC data requires not only technical knowledge of the complicated schema of the IFC documentation but also an experience in manipulating the different BIM software to extract the required information from the building model data. Moreover, many IFC open-source parsing tools can lose the semantics of IFC when different entities are combined or compared across IFC files (Zhao et al., 2020).

Several research studies attempted to overcome these challenges by introducing automatic methods and tools to facilitate the parsing of complex data from IFC files. These studies mainly focused on transforming IFC files into an alternative data format via mapping rules (Kang and Hong, 2018), graph-theory-based technology (Zhu et al., 2023), a structured query language (Krijnen and Beetz, 2018) or an ontology-based language (Liu et al., 2016). However, in such methods, an in-depth knowledge of query formulation and mapping rules is required in order to extract an attribute or an entity relation. Moreover, when the rules are incomplete, the extracted attribute may also be incomplete or irrelevant to the user requirements (Du et al., 2021).

Recently, a number of studies have adopted Natural Language Processing (NLP) techniques to provide natural language-based information extraction from IFC data. NLP-based information retrieval systems have focused on parsing IFC data via syntactic tagging tools (Park and Kang, 2018), dependency rules (Wang and Issa, 2019), and keyword classification (Wu et al., 2019). Although, the existing effort to parse IFC data in natural language queries and responses is limited, it proved to significantly facilitate the process of information acquisition from building information models (Lin et al., 2016; Wang et al., 2022). Such query-answer (QA) models parsing IFC in natural language directly extract information from BIM data without complex reasoning and expensive computation.

In this research, we advance NLP methodologies by harnessing the power of generative AI and Graph-RAG techniques to create a QA system capable of processing IFC data where both queries and responses are constructed in plain language. Graph-RAG is a specialized variant of the Retrieval-Augmented Generation (RAG) framework that incorporates graph structures into the retrieval and reasoning process. We show that Graph-RAG offers an effective approach for extracting and processing IFC data for two main reasons. First, the intricate relationships between building components in IFC standards such as walls, doors, floors, and their properties are naturally graph-like: nodes represent entities like building elements (e.g., *IfcWall*, *IfcDoor*) or attributes (e.g., Material, LoadBearing) and edges represent relationships between these entities, such as spatial containment (*IfcRelContained*, *InSpatialStructure*), or connections (*IfcRelConnects*) (buildingSMART International Standards, 2019). Second, instead of retrieving isolated pieces of data, Graph-RAG retrieves sub-graphs or interconnected entities relevant to a query. For example, for a query like *"What materials are used in the walls of the second floor?"* Graph-RAG retrieves the sub graph linking the second-floor spatial structure, its contained walls, and their materials. This ensures that the system retrieves not just raw data but also the necessary contextual information, enabling accurate answers.

As shown in Figure 1, our Graph-RAG retrieval system queries a generative language model, specifically GPT-4o (Hurst et al., 2024), by asking questions in natural language about specific information in the the IFC data. For example, the answer to the input query in our question test set:'*What is the thermal transmittance of the glass pan-* els?" is extracted from the IFC graph data which enhances the GPT-4O retriever with the graph path relevant to the question. The LLM uses both the query enhancement and graph-knowledge enhancement to give the correct response in natural language – '*The thermal transmittance of the glass panels is 6.7069*'. Thus, to extract specific information from the IFC model, the system uses the retrieved sub-graphs or nodes of the IFC data to augment the query input to GPT-4o; the latter utilises its pretrained knowledge base to translate the graph response into a natural language answer. By incorporating structured relational data, the language model gains a deeper understanding of the context (with 68% accuracy), reducing issues like hallucination or inconsistencies. Moreover, since IFC data is complex and often difficult to interpret for non-technical stakeholders, our Graph-RAG approach bridges this gap by allowing natural language queries and generating responses in plain language. We explain in the Methodology section, how we construct the Graph-RAG retrieval model to answer queries, provide explanations, and return responses on the relationships within the IFC graph.

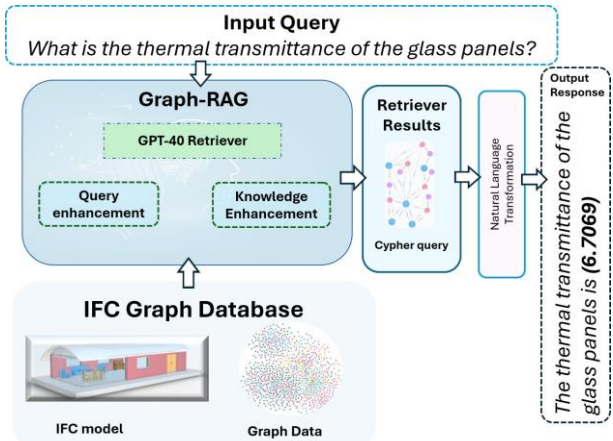

*Figure 1: Overview of the Graph-RAG framework for parsing the IFC data with a natural language QA system.*

To demonstrate our experiments, the paper first presents a brief review of the major research conducted for parsing IFC data as well as recent advances in NLP research concerned with utilising generative language models and the Graph-RAG technique in information retrieval. Next, the Methodology section outlines our two-stage approach for graph generation and Graph-RAG-based information retrieval from IFC data. Experimental results are presented in the Experiments and Results section. And finally, the Discussion and Conclusion sections address the strengths, limitations, and future directions of the proposed approach. Both the IFC data and code for the LLM-assisted Graph-RAG system are available at (link obliterated for anonymity).

## Related Research

### IFC Parsing

Research on information extraction from IFC models has been conducted for various purposes, such as model compliance checking (Zhang and El-Gohary, 2016), IFC file comparison (Shi et al., 2018), data retrieval, interpretability, and integration (Du et al., 2021). Studies on information extraction from IFC files have focused on retrieving accurate and complete information from different IFC elements. Typically, IFC attributes are accessed by mapping the file to a different data format, where information can be extracted via algorithm-based approaches (Wong Chong and Zhang, 2024), the use of graph theory and ontology-based models (Zhu et al., 2023), or structured query languages such as SPARQL (Xinglei et al., 2022). However, these methods require technical understanding of the query language, such as SPARQL, and experience with the target data format into which the IFC data is mapped.

To overcome the complexity of mapping-based IFC retrieval methods, there is growing interest in applying NLP techniques for extracting information from IFC data. For example, Lin et al. (2016) develops an intelligent data retrieval and representation approach for cloud BIM applications, using NLP techniques to map user requirements to IFC entities and properties via keyword and constraint classification. Wang et al. (2022) uses NLP methods to develop a three-stage QA system for retrieving information from IFC data. Their system consists of a natural language query analyser, a content word classifier, and a language generation model that transforms retrieved BIM information into natural language output.

Most NLP-based studies aim to represent both queries and answers of information extraction system in plain language (Wang and Issa, 2020). However, these systems usually rely on complex NLP pipelines, including POS tagging, dependency parsing(Wang et al., 2022), syntactic tree generation(Wu et al., 2019) or machine learning classifiers to link regulatory and IFC concepts (Zhang and El-Gohary, 2023). Few studies deployed generative LLMs for the information retrieval of BIM data. Zheng and Fischer (2023) introduced a prompt-based virtual assistant framework using generative LLM (GPT) to support NLP-based BIM information retrieval. To the best of our knowledge, our proposed method- harnessing the power of generative AI and Graph-RAG techniques to create a QA system capable of processing IFC data- has not yet been explored. The following section outlines how Graph-RAG enhances information retrieval by integrating external graph-based knowledge to address LLM limitations.

### Graph Retrieval-Augmented Generation

Despite their advanced language comprehension and text generation capabilities, Large Language Models (LLMs) face limitations due to gaps in domain-specific knowledge, real-time updates, and proprietary information beyond their pre-training corpus. These gaps often lead to ''hallucination'', where inaccurate or fabricated content is generated. Retrieval-Augmented Generation (RAG), is an information retrieval technique that was initially introduced by Meta AI (Lewis et al., 2020) to improve the LLMs performance to retrieve correct answers from textual data. The basic idea of RAG is that both the query and text chunks from the dataset are encoded into vectors using the same LLM (e.g Llama 3 (Dubey et al., 2024) GPT-4 (Achiam et al., 2023), Qwen2 (Chu et al., 2024), etc.). Encoding the entire dataset provides the retrieval model additional context, thus improving its ability to find accurate answers. The RAG technique acts like a search engine where a retrieval model checks for the most relevant text chunk(s) in the dataset vector space based on the query vector, then the second component of the model, the LLM, provides the retrieved answer in natural language. By utilising an external knowledge base, RAG has proven highly effective in enhancing LLM outputs as it addresses problems such as "hallucination" caused by outdated information or insufficient domain-specific knowledge not included in the LLMs' pre-training data (Siriwardhana et al., 2023; Peng et al., 2024; Fan et al., 2024).

Graph Retrieval-Augmented Generation (Graph-RAG) is an advanced extension of the RAG framework that integrates graph-based knowledge structures to improve information retrieval and reasoning in LLMs (Guan et al., 2024). Unlike traditional RAG, which retrieves unstructured text, Graph-RAG retrieves structured graph elements (such as nodes, triples, paths, or sub-graphs) to enhance context-aware and relational knowledge generation. By retrieving sub-graphs or graph communities, Graph-RAG enhances contextual understanding, making it particularly effective for query-focused tasks. Since user queries are typically in natural language and LLMs excel at natural language comprehension, a key strategy is translating retrieved graph data into descriptive, user-friendly text.

Various methods have been proposed for information retrieval using the Graph-RAG, such as predefined natural language templates for graph edges (Ye et al., 2023), structured descriptions of neighbouring nodes (Edge et al., 2024), and LLM-generated summaries for graph communities (Huang et al., 2019). Due to its success in information retrieval problems with complex data, Graph-RAG is now applied not only in downstream NLP tasks, but also in practical domains such as e-commerce (Afrose, 2024), biomedical (Wu et al., 2024), and law (Phyu et al., 2024), which reflects its versatility and adaptability across sectors. In our use case, we utilise Graph-RAG to build a knowledge graph from IFC data to reinforce the LLMs retrieval capability to fetch relevant sub-graphs containing contextual and relational information pertinent to a user query about the building model. Moreover, since queries are in natural language, the Graph-RAG technique is used to translate structured IFC graph data into text via few-shot prompting of the LLM model which synthesises the retrieved IFC graph-based information into a coherent and a contextually rich response. Details of our methodology are provided in the next section.

## Methodology

The IFC interpretation method involves two stages: transforming the IFC model into a graph, and using an LLM to interpret the graph and answer user queries. This section details each stage.

**Preliminaries: Industry Foundation Classes**

An IFC file consists of two parts: a Header, containing metadata like creation date and schema version, and a Data section, which holds the core building model, including geometric and semantic details and entity relationships. Figure 2a provides an example that illustrates the structure of an IFC file. Each entry starts with an entity ID (e.g. #29), followed by the entity name (e.g. *IfcWall*), and its properties (e.g. '*Graphisoft*' ). Relationships are represented by reference IDs (e.g. #28, #5). Some entities include a Global Unique Identifier (GUID) for distinct identification (e.g. Global Id in Figure 2a). However, in BIM applications, GUIDs frequently change when the file is exported Zhao et al. (2020).

Interpreting IFC structures is challenging due to their hierarchical complexity. This complexity stems not only from the dense web of relationships common in building projects but also from the lack of explicit property and reference names in IFC entities. As a result, users unfamiliar with the schema often struggle to extract meaningful information. This challenge is shown in Figure 2a, where a user without prior knowledge of IFC may struggle to understand the value '*Haustuer*' represents the '*Name*' property in the IFC entity #7060. This highlights the need for tools that simplify IFC data extraction and interpretation to improve accessibility and usability.

**Stage 1: Graph Generation**

To systematically represent the extracted IFC data as a graph, we define a structured approach for node and edge creation. Each entity within the IFC file is represented as a node (**n**). An edge (**e**) is defined between two nodes, **n** and **m**, if the property list of **m** contains a reference identifier pointing to **n**, and vice versa. For the sake of simplicity in graph construction and query generation in next steps, all edges are considered undirected. To construct the graph from an IFC file, we utilised the IFCOpenShell library in Python (IFCOpenShell, 2024). IFCOpenShell is an open-source toolkit designed for parsing, editing, and processing IFC files. This library simplifies efficient traversal of IFC entities by identifying entities, their attributes, and their relationships. Using IFCOpenShell, we traversed all entities in the IFC file in ascending order based on their entity IDs. For each entity, we defined:

- The entity name as the node label (e.g. *IFCDoor*).
- The entity ID as a node id feature (e.g. #7060).
- The set of property names and values as node attributes (e.g. *Name:Haustuer*).

Similarly, for each reference ID that exists in the property set of an entity, we define an edge labelled with the corre-sponding property name. This label further specifies the type of relationship in our graph. Figure 2b illustrates a sample graph generated from the IFC file presented in Figure 2a.

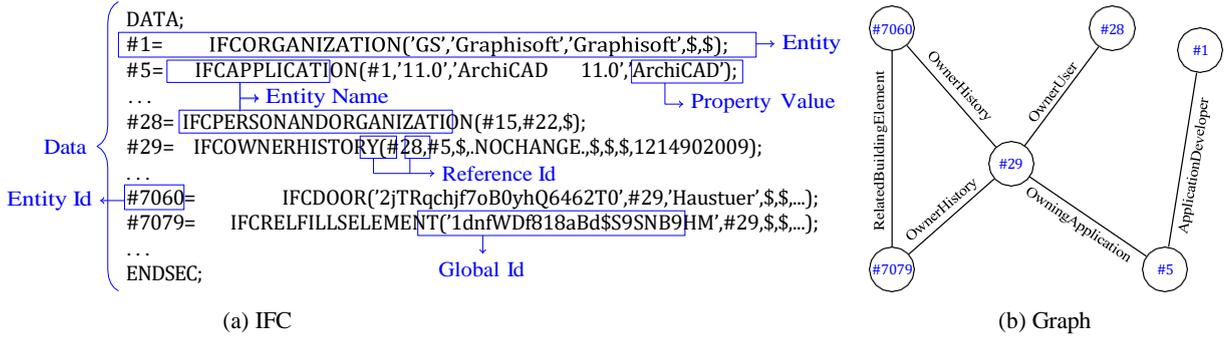

Figure 2: (a) A sample Data section of an IFC file, and (b) A sample graph generated from the entities shown in the sample IFC file

From the graph, we can observe how edges are defined between entities (nodes). With each edge label showing the name of the related property. For instance in Figure 2b, the edge label between nodes **#1** and **#5**, corresponds to the name of the first property in entity #5, which is *ApplicationDeveloper*.

Finally, the extracted graph data is organized into separate node and edge lists, providing a structured format that enables efficient querying and seamless integration into subsequent processing steps within the methodology.

### Stage 2: GraphRAG-Based Query Interpretation and Response Generation

In our designed system, the LLM provides a natural language interface to the IFC graph database which we query with a cypher query language. Cypher is a declarative graph query language that allows for expressive and efficient data querying in a property graph used primarily in Neo4j graph database (Neo4j, 2024). To enhance Cypher query generation, we construct a prompt-based system guiding the LLM to use only the provided IFC schema which is constructed in the graph generation stage. The prompt included:

1. Schema constraints (*e.g.*, ensuring correct relationships between entities).
2. Specific formatting rules (*e.g.*, not using meaningful variables for cypher).
3. Few-shot learning examples with correct Cypher queries and responses.

We used the few-shot learning approach in order to refine the LLM query generation task. Few-shot prompting is a technique used with generative LLMs where the model is provided with a few examples (shots) of input-output pairs before generating a response. This helps the model understand the task better without requiring fine-tuning. Accordingly, we provided our retriever model with structured examples that demonstrated: 1) a question, 2) correct graph database query cypher, 3)query result, and 4) expected answer. Table 1 illustrates sample few-shot learning examples used to guide the LLM. By including diverse examples, the model is guided towards the correct query patterns. This method also helped mitigate common errors such as using incorrect edge labels or retrieving attributes from the wrong IFC entities, minimising hallucinations or inaccurate responses.

To further optimize the LLM-driven query interpretation, we iteratively refined the prompt instructions based on empirical results. Through multiple test iterations, we identified cases where the model struggled with ambiguous queries such as distinguishing between *Roof* as an IfcSpace (a spatial unit like a room) and *Roof* as the IfcRoof (a structural element). So that additional clarification rules were introduced into the prompt, ensuring that the LLM correctly mapped user intent to the corresponding IFC elements.

Following query generation, the retrieved results were processed into natural language responses to ensure clarity for end users. Hence, instead of returning raw data structures from query results, the LLM converts the IFC query outputs into human-readable answers. This approach helps connect structured database searches with easy-to-understand responses which makes the system easier to use for people who are not familiar with the complex structure of the IFC or graph-based query languages.

### Experiments & Results

For our experiments, we selected a sample IFC data file from the buildingSMART Community Projects(buildingSMART Community, 2024).The selected IFC file represents a small-scale building model, as illustrated in Figure 3a.

The selected IFC model was converted into a graph-based representation, as explained in the methodology section in Stage 1. The produced graph from this IFC file contains 47310 nodes and 82652 edges, partly visualized in Figure 3b. The extracted node and edge were then imported into a Neo4j database using Neo4j Desktop (version 5.12.0). We utilised the Neo4j platform for this stage because of its ability to facilitate efficient relationship-based retrieval and supports complex IFC queries beyond simple key-value searches.

For the IFC interpretation and question-answering system, we integrated the OpenAI API within a Python en-vironment, and the GPT4o as our baseline LLM for our benchmark. The LLM was responsible for translating natural language questions into Cypher queries and generating human-readable responses based on the extracted results. For the prompt construction and query execution, we utilised LangChain(LangChain, 2024), which is a framework designed for managing LLM-based tasks. LangChain was used to dynamically structure the prompt and to ensure that the LLM received only the relevant IFC schema and relationship constraints. Additionally, LangChain facilitated post-processing of Cypher queries outputs by converting structured results into natural language responses.

*Table 1: Samples of Fewshot examples provided to LLM*

| | |
|---|---|
| Question | How many rooms exist in this building? |
| Cypher | MATCH (n:IfcSpace) RETURN COUNT(n) AS RoomCount |
| Context | {'RoomCount': 4 } |
| Response | There are 4 rooms in the building. |
| Question | What is the level of the living room? |
| Cypher | MATCH (n1:IfcSpace)-[r1:RelatedObjects]-(n2:IfcRelDefinesByProperties)-[r2]-(n3:IfcPropertySet)-[r3]-(n4:IfcPropertySingleValue) WHERE toLower(n1.Name) CONTAINS toLower("Living room") and toLower(n4.Name) CONTAINS toLower("Level") RETURN n4.NominalValue As Level |
| Context | {'Level ': IfcLabel('Level: Ground Floor') } |
| Response | Living room level is Ground Floor. |
| Question | What is the NetFloorArea of the living room? |
| Cypher | MATCH (n1:IfcWall)-[r1]-(n2:IfcRelDefinesByProperties)-[r2]-(n3:IfcPropertySet)-[r3]-(n4:IfcPropertySingleValue) WHERE toLower(n4.Name) CONTAINS toLower("Height") RETURN n4.NominalValue |
| Context | {'n4.AreaValue': "51.9948250000001" } |
| Response | The net floor area of living room is 51.99" |

*Table 2: Examples of the user queries and GraphRAG responses*

| User Question | Expected Response | Cypher Query | LLM Response |
|---|---|---|---|
| Q1. How many doors exist in the building? | There are 3 doors. | MATCH (n:IfcDoor) RETURN COUNT(n) AS DoorCount | There are 3 doors in the building. |
| Q2. What is the volume of the roof space? | The gross volume of the roof space is 76.47 cubic meters. | MATCH (n1:IfcSpace)-[r1:RelatedObjects]-(n2:IfcRelDefinesByProperties)-[r2:RelatingPropertyDefinition]-(n3:IfcElementQuantity)-[r3:Quantities]-(n4:IfcQuantityVolume) WHERE n1.Name CONTAINS "Roof" AND n4.Name CONTAINS "Volume" RETURN n4.VolumeValue | The volume of the roof space is 76.46551559765. |
| Q3. How many storey exist in the building? | There are 2 storey in the building | MATCH (n:IfcBuildingStorey) RETURN COUNT(n) AS StoreyCount | The building has 2 storeys. |
| Q4. What is the gross floor area of the entrance hall? | The gross floor area of the entrance hall is 8.69 square meters. | MATCH (n1:IfcSpace)-[r1:RelatedObjects]-(n2:IfcRelDefinesByProperties)-[r2]-(n3:IfcElementQuantity)-[r3]-(n4:IfcQuantityArea) WHERE n1.Name CONTAINS "Entrance hall" AND n4.Name CONTAINS "GrossFloorArea" RETURN n4.AreaValue | The gross floor area of the entrance hall is 8.69350624999999. |
| Q5. What is the illuminance unit defined in the file? | The illuminance unit defined in the file is Lux. | MATCH (n1:IfcSIUnit) WHERE n1.UnitType = "ILLUMINANCEUNIT" RETURN n1.Name ['n1.Name': 'LUX'] | The illuminance unit defined in the file is LUX. |
| Q6. Is there a Laundry in the building? | No there isn't any Laundry space in the building. | MATCH (n1:IfcSpace) WHERE toLower(n1.Name) CONTAINS toLower("Laundry") RETURN COUNT(n1) > 0 AS IsLaundryPresent | There is no Laundry in the building. |
| Q7. What is the perimeter of the entrance hall? | The entrance hall has a perimeter of 12,810. | MATCH (n1:IfcSpace)-[r1:RelatedObjects]-(n2:IfcRelDefinesByProperties)-[r2]-(n3:IfcElementQuantity)-[r3]-(n4) WHERE toLower(n1.Name) CONTAINS toLower("entrance hall") AND toLower(n4.Name) CONTAINS toLower("perimeter") RETURN n4.LengthValue | The perimeter of the entrance hall is 12,810.0. |
| Q8. What is the building address? | The address is 'Westminster, London, UK'. | MATCH (n1:IfcBuilding)-[r1:BuildingAddress]->(n2:IfcPostalAddress) RETURN n2.AddressLines, n2.Town, n2.Region, n2.PostalCode, n2.Country | Enter address here, Westminster, London, UK. |
| Q9. What is the unconnected height of the roof space? | The roof has an unconnected height of 1000 mm (1 meter). | MATCH (n1:IfcSpace)-[r1:RelatedObjects]-(n2:IfcRelDefinesByProperties)-[r2]-(n3:IfcPropertySet)-[r3]-(n4:IfcPr WHERE toLower(n1.Name) CONTAINS toLower("roof") AND toLower(n4.Name) CONTAINS toLower("Height") RETURN n4.NominalValue AS RoofSpaceHeight | The unconnected height of the roof space is IfcLengthMeasure(0.) and IfcLengthMeasure(1000.). |
| Q10. What is the name of the project? | The project name is 'Project Name'. | MATCH (n1:IfcProject) RETURN n1.Name AS ProjectName | I don't know the answer. |

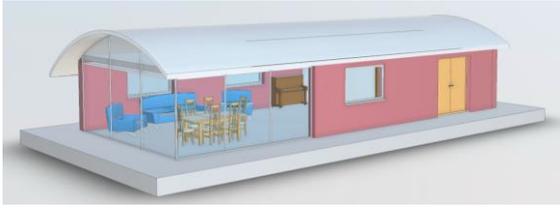

(a) IFC Model

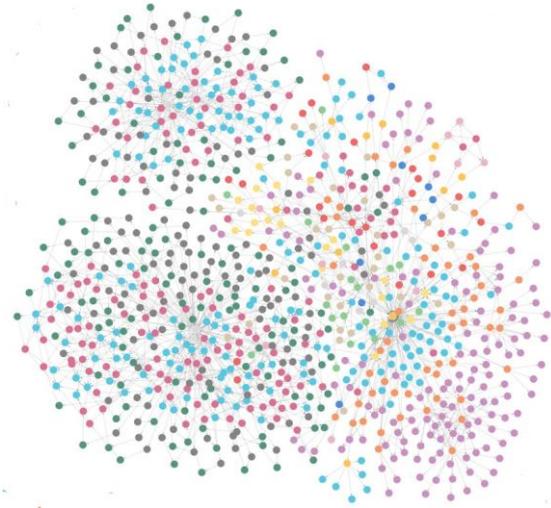

(b) Generated Graph

*Figure 3: (a) The IFC Building model used in experiments, and (b) The generated graph from IFC Building model*

We compiles a benchmark dataset of 60 question-answer pairs, serving as ground truth for evaluating the model's performance. The question selection process focused on covering a range of complexities, from simple node property retrieval to more advanced queries requiring traversal of multiple relationships in the IFC graph. In other words, our evaluation tests the model's ability to construct and execute various types of Neo4j Cypher queries, as well as its ability to generate accurate and coherent natural language responses based on the results of the queries. This evaluation tested the model's ability to:

- Generate syntactically and semantically correct Cypher queries.
- Retrieve accurate results from the Neo4j IFC graph.
- Convert retrieved values into natural language responses.

Table 2 provides a sample of the tested questions, their expected outputs, the Cypher queries as well as the natural language responses generated by LLM.

The results indicate that the model performs well on queries, involving entity counts. For instance, questions such as "How many doors exist in the building?" yielded accurate results, demonstrating the model's ability to correctly interpret and execute queries involving count-based retrieval. Similarly, numerical attributes such as "What is the volume of the roof space?" and "What is the gross floor area of the entrance hall?" were successfully extracted. However, a minor issue was observed in the formatting of numerical outputs (e.g., Q4), where floating-point precision resulted in values being displayed with an excessive number of decimal places (e.g., showing 14 digits after the decimal point) instead of a properly rounded value like 8.69. Another observation relates to the system's ability to process semantic variations in queries. The response to Q6 correctly identified that no laundry space was present, showing that the model can handle boolean-type queries effectively.

While the system performed well on structured queries, as it can be seen in the table, certain responses, such as retrieving the project name (Q9), resulted in failure cases where the LLM returned "I don't know the answer", despite the correct Cypher query execution. These cases highlight areas for further optimization in the future.

## Discussion

This study explored the application of Graph-RAG for parsing IFC files in natural language using a prompt-based approach. By leveraging GPT-4o as the retriever model, we aimed to enhance the accessibility of IFC data for users without requiring specialized domain knowledge or complex query languages. Our results demonstrate that the proposed system effectively extracts and retrieves information from the IFC graph model, enabling intuitive querying of BIM data. However, several challenges emerged in handling more intricate IFC structures, which highlights areas for further improvement.

Despite its effectiveness in simpler queries, the system exhibited limitations in handling complex graph paths and multiple entity references. When queries required traversing multiple relationships across the IFC graph (e.g., retrieving materials used in a wall while also linking to its load-bearing properties), the retriever often struggled to identify the correct traversal sequence, leading to incomplete or inaccurate responses. This suggests that the model's graph reasoning capabilities require further refinement, possibly through graph-based reinforcement learning or enhanced graph traversal heuristics. Moreover, when an entity was referenced multiple times in different contexts (e.g., Roof provided as a space and a sepa-rate building element), GPT-4o sometimes failed to disambiguate the references correctly. This resulted in conflicting or generalized responses, indicating a need for improved entity disambiguation techniques. One potential solution is to augment the retrieval mechanism with additional context encoding strategies, such as hierarchical retrieval or entity linking methods. Finally, as the system relied on a prompt-based approach, query formulation significantly impacted retrieval performance. In some cases, the model failed to give the natural language answer despite the fact that it managed to produce the correct cypher query. This underscores the importance of refining prompt engineering techniques to guide the model towards more accurate interpretations of IFC data.

## Conclusions and Future Work

Overall, this study demonstrates the potential of GraphRAG for parsing IFC data in a user-friendly, natural language format. Future work will include experiments with multiple building models of varying types and sizes, as well as a more diverse and extensive set of question-answer pairs, to evaluate the robustness and generalizability of our retrieval approach.

While GPT-4o successfully retrieves relevant information, its performance degrades with complex graph paths and multiple entity references. Our future research will focus on enhancing graph traversal techniques, by implementing path-ranking algorithms or query-dependent traversal methods to improve retrieval over complex IFC structures. To mitigate the complexity and deeply nested hierarchy of IFC files, we plan to investigate focus-based methods that isolate and process specific parts of a building model, allowing more targeted and efficient retrieval. We will also explore the possibility of integrating Named Entity Recognition (NER) and reference resolution techniques to handle multiple references more effectively to make graph-based IFC retrieval more reliable and scalable for real-world BIM applications. Additionally, the development of a dedicated user interface tool will further enhance usability and integration with existing BIM workflows.